\documentclass{article}

\usepackage[final, nonatbib]{neurips_2023}

\usepackage[utf8]{inputenc} 
\usepackage[T1]{fontenc}    
\usepackage{hyperref}       
\usepackage{url}            
\usepackage{booktabs}       
\usepackage{amsfonts}       
\usepackage{nicefrac}       
\usepackage{microtype}      
\usepackage{xcolor}         
\usepackage{amsmath}        
\usepackage{graphicx, wrapfig, svg}
\usepackage{caption}
\usepackage{subcaption}
\usepackage{bm}
\usepackage[sorting=none]{biblatex}

\usepackage[colorinlistoftodos]{todonotes}

\usepackage{color}

\newcommand{\bp}{\bm{p}} 
\newcommand{\bxi}{\bm{\xi}}
\newcommand{\bx}{\bm{x}}
\newcommand{\bX}{\bm{X}}

\title{Exploring the Temperature-Dependent\\ Phase Transition in Modern Hopfield Networks}

\author{%
  Felix Koulischer \qquad Cédric Goemaere \qquad Tom Van Der Meersch \\ 
  \textbf{Johannes Deleu \qquad Thomas Demeester}\\ 
  IDLab, Ghent University -- imec\\
  Ghent, Belgium \\
  \texttt{first.last@ugent.be} \\
}

\begin{document}

\maketitle

\begin{abstract}
    The recent discovery of a connection between Transformers and Modern Hopfield Networks (MHNs) has reignited the study of neural networks from a physical energy-based perspective. This paper focuses on the pivotal effect of the inverse temperature hyperparameter $\beta$ on the distribution of energy minima of the MHN. To achieve this, the distribution of energy minima is tracked in a simplified MHN in which equidistant normalised patterns are stored. This network demonstrates a phase transition at a critical temperature $\beta_{\text{c}}$, from a single global attractor towards highly pattern specific minima as $\beta$ is increased. Importantly, the dynamics are not solely governed by the hyperparameter $\beta$ but are instead determined by an effective inverse temperature $\beta_{\text{eff}}$ which also depends on the distribution and size of the stored patterns. Recognizing the role of hyperparameters in the MHN could, in the future, aid researchers in the domain of Transformers to optimise their initial choices, potentially reducing the necessity for time and energy expensive hyperparameter fine-tuning.
\end{abstract}

\section{Introduction}
Energy based models (EBMs) have been around for over twenty years, but were until recently regarded by many as historical artefacts, due to overwhelming successes in deep neural networks trained with backpropagation \cite{Goodfellow2016_DL}. However, recent developments have once again brought EBMs into the spotlight \cite{NatureReviews_Krotov}, aligning with the surging interest in more energy-efficient \cite{ZoppoMarrone2020} or biologically plausible neural networks \cite{Bengio2015_biologicalDL}. The main constraint of the prototypical EBM, the Hopfield Network (HN) \cite{OriginalHopfieldNetworks_1982, ContHN_1984_LimitedStorageCapacity}, namely its low storage capacity, was remedied by increasing the interaction order inside the hamiltonian \cite{DAMs_Krotov_Hopfield_1, DAMs_Krotov_Hopfield_2}. Such networks, known as Dense Associative Memory models (DAMs), possess a polynomial storage capacity $C$, i.e. with $d$ the dimension of states within the network and $n$ the interaction order, one has that \(C \propto d^{n-1}\). Going a step further and using an exponential interaction, effectively considering an infinitely large interaction order, an exponential storage capacity was obtained, 
i.e. \(C \propto \alpha^d\) where $\alpha$ is a constant \cite{ExpOrderInt_Demercigil, HopfieldNetworksIsAllYouNeed, PreciseExponentialStorageCapacity_Lucibello}. Interestingly, the EBM consisting of such an exponential interaction vertex, referred to as the Modern Hopfield Network (MHN), displays a strong connection to a widespread neural network architecture, the Transformer \cite{HopfieldNetworksIsAllYouNeed, AttentionIsAllYouNeed}. The connection between the two arises from the specific mathematical form of the energy minimising update rule, consisting of a Softmax averaging operation reminiscent of the attention mechanism present in Transformers. While the Transformer has become the go-to architecture in many fields \cite{TransformersSucces}, the understanding of its inner workings remains limited \cite{EnergyTransformer}. This is in stark contrast with the MHN, which is very well described from a theoretical point of view, while currently remaining rarely used in practice \cite{HopfieldNetworksIsAllYouNeed, EnergyTransformer}. It is hoped that the theoretical understanding of MHNs could help guide research towards novel Transformer architectures \cite{EnergyTransformer}.\\ 
The energy function of the MHN, whose expression strongly resembles the Helmholtz free energy in the canonical ensemble from statistical physics, is very sensitive to the inverse temperature hyperparameter $\beta$ (sometimes referred to as the interaction strength \cite{PreciseExponentialStorageCapacity_Lucibello}). Amongst other, $\beta$ determines the peakedness of the arising Softmax function by rescaling the dot-product inside the exponential interaction. It is also known that $\beta$ has an impact on the distribution of the minima of the energy function \cite{HopfieldNetworksIsAllYouNeed, PreciseExponentialStorageCapacity_Lucibello}. At large $\beta$, the minima reside close to single stored patterns, while at lower values the minima often reside in the linear combination of multiple similar patterns \cite{HopfieldNetworksIsAllYouNeed}. This paper, akin to the approach taken by C. Lucibello and M. Mézard \cite{PreciseExponentialStorageCapacity_Lucibello}, investigates the transformation of the distribution of the energy minima of the MHN as $\beta$ varies from zero to infinity. At very high temperatures (\(\beta \approx 0\)) the energy landscape is maximally blurred, and a single global minimum is present at the origin \cite{ HopfieldNetworksIsAllYouNeed, PreciseExponentialStorageCapacity_Lucibello}. Conversly, at very low temperatures (\(\beta \approx \infty\)) the energy landscape consists of many sharp minima residing in each of the stored patterns. 
Between these two limiting scenarios, a certain phase transition occurs. In contrast to C. Lucibello and M.Mézard's work, we will assume the patterns stored in the MHN to be equidistant and normalized, rather than Gaussian or uniformly distributed spherical vectors. This particular simplifying assumption allows for an analytical treatment, which yields useful insights into MHNs'
inner workings. Specifically, it allows computing the critical temperature $\beta_{\text{c}}$ at which the phase transition occurs. Crucially, it appears that the network dynamics are governed by a physical quantity we will refer to as the \emph{effective inverse temperature}, denoted as $\beta_{\text{eff}}$. 
We will show how $\beta_{\text{eff}}$ depends on the hyperparameter $\beta$, as well as on the distribution and size of the stored patterns.\\
Understanding how $\beta$ affects the network dynamics could guide researchers working on MHNs, and potentially even on Transformers, towards better hyperparameter selection, which is known to be an energy and time consuming aspect to practical neural networks training \cite{EnergyConsiderationsDNNs}. 
\section{Theoretical considerations on phase transitions in MHNs}
\textbf{Modern Hopfield Networks}\newline
Primarily designed to work as an associative memory model, the MHN stores $N$ patterns $\mathbf{x_i}$ of dimension $d$. These patterns are grouped together as the columns of a single matrix $\mathbf{X}$. As all EBMs, its dynamics are governed by the minimisation of its energy function in the state space represented by the state vector \(\bm{\xi} \in\mathbb{R}^d\). The considered energy function is given by:
\begin{equation}
E\left(\bm{\xi},\mathbf{X},\beta\right) = -\frac{1}{\beta} \text{ln}\sum_{i=1}^N\exp\left(\beta \mathbf{x_i}^T\bm{\xi}\right)+\frac{1}{2} \bm{\xi}^T\bm{\xi}
\label{eq: Energy MHN}
\end{equation}
Application of the concave-convex procedure (CCCP) \cite{LitStud_CCCP} leads to the following energy minimising state update rule that expresses the subsequent state $\bxi'$ as a function of the current state $\bxi$ and the given patterns $\bX$:
\begin{equation}
   \bm{\xi}' = \mathbf{X}\bp,\quad\text{with}\;\;  \bp = \frac{\exp\big(\beta\bX^T\bxi\big)}{\sum_{k=1}^N\exp\big(\beta\bx_k^T\bxi\big)}
\label{eq: Modified update eq}
\end{equation}
%
Here, the stochastic probability $\bm{p}$-vector has been introduced, containing the recognizable Softmax operator. It is clear that as the state $\bm{\xi}$ converges towards a stable state \(\bm{\xi}^{\ast}\), the $\mathbf{p}$-vector also converges towards a stable state \(\mathbf{p}^{\ast}\)\footnote{Technically, this is only correct up to the kernel of the matrix $\mathbf{X}$ (i.e. \(\bm{\xi}' = \bm{\xi} \iff \mathbf{p}' = \mathbf{p} +\text{ker}(\mathbf{X})\)). It was however verified that as $\mathbf{\xi}$ converged towards its equilibrium value, the same held for $\mathbf{p}$.}. It is therefore sufficient to look at how the $\mathbf{p}$-vector changes under repeated application of the Softmax update rule to understand how $\beta$ affects the equilibration dynamics. Moreover, the $\mathbf{p}$-vector serves as a direct representation of the proximity of a given state to the stored patterns, making it a valuable tool to understand how the distribution of energy minima depends on the positions of these patterns.

\textbf{Effective inverse temperature}\newline
In order to facilitate the analytical analysis of the role of the inverse temperature $\beta$, the following simplifying assumptions on the nature of the stored patterns $\bx_i$ are made. 
First, the stored patterns are assumed to have identical norms, i.e. \(\left \| \mathbf{x_i} \right \|= \left \| \mathbf{x} \right \|\text{, } \forall i=1,..,N\). Second, it is assumed that all patterns have the same separation angle between them, i.e. \(\boldsymbol{x_i}^T \boldsymbol{x_j} = \left \| \boldsymbol{x} \right \|^2\cos(\theta) \text{, } \forall i\neq j\), with $\theta$ the angle between both patterns\footnote{This imposes a restriction on the number of stored patterns $N$, namely \(N \leq d+1\).}.
Geometrically, the stored patterns can be thought of as equidistant points on the $d$-sphere.  
The above assumptions can be expressed as
\begin{equation}
    \big(\bX^T\bX\big)_{ij} = \|\bx\|^2\big(\cos\theta + (1-\cos\theta)\delta_{ij}\big)
    \label{eq:XTX}
\end{equation}
with $\delta_{ij}$ the Kronecker delta.
For the constrained patterns satisfying \eqref{eq:XTX} we find, by substitution of the state update equation~\eqref{eq: Modified update eq},
\begin{equation}
\bX^T\bxi' = \bX^T\bX\bp = \|\bx\|^2\big( \cos\theta\,\mathbf{1} + (1-\cos\theta)\,\bp\big) 
\end{equation}
such that the update equation for $\bp$ can be written as
\begin{equation}
    \bp' = \frac{\exp\big(\beta\bX^T\bxi'\big)}{\sum_{k=1}^N\exp\big(\beta \bx_k^T\bxi'\big)}
    = \frac{\exp\big(\beta\|\bx\|^2(1-\cos\theta)\,\bp\big)}{\sum_{k=1}^N\exp\big(\beta\|\bx\|^2(1-\cos\theta)\,p_k\big)}
    = \frac{\exp\big(\beta_\text{eff}\,\bp\big)}{\sum_{k=1}^N\exp\big(\beta_\text{eff}\,p_k\big)}
    \label{eq: update rule for p}
\end{equation}
with the introduction of the \emph{effective} inverse temperature $\beta_\text{eff} = \beta \left \|\mathbf{x}  \right \|^2 \left(1- \cos(\theta)\right)$, 
which can be seen as the relevant parameter that directly governs the equilibrium dynamics. Due to the simplicity of the pattern correlations, the MHN dynamics mimic those of the iterated softmax \cite{TinoISM2, TinoISM}.

The obtained expression for $\beta_\text{eff}$ implies that when the stored patterns are highly similar (\(\cos(\theta)\rightarrow 1\)), the hyperparameter $\beta$ has to be chosen significantly higher to exhibit the same dynamics (i.e., a similar $\beta_\text{eff}$) as a network consisting of markedly distinct patterns. As explained by Ramsdauer et al., when multiple patterns are in close proximity they tend to form an energy minimum lying close to their mean, referred to as metastable state \cite{HopfieldNetworksIsAllYouNeed}. To further enhance a pattern specific memory, it makes sense that $\beta$ should be increased further. The dependence of $\beta_{\text{eff}}$ on the patterns norms stems from the quadratic scaling of the dot-product on the former. Crucially, this implies that distinct $\beta$ values are required to obtain similar dynamics depending on whether normalisation is used in the network.

\textbf{Phase transition and critical inverse temperature}\newline
Another important note is that the uniform $\mathbf{p}$-vector, defined by \(p_i = \frac{1}{N}\text{ , }\forall i=1,.., N\), is always a fixed point of the update rule defined by \eqref{eq: update rule for p}. This fixed point further corresponds to the high temperature limit, as \(\beta_{\text{eff}}\rightarrow0\) one has that \(p_i \rightarrow\frac{1}{N}\). To understand the low temperature limit, it is useful to consider that the $\mathbf{p}$-vector has one component, say $p_I$, that is larger than all others, i.e. \(p_I > p_i \text{ , } \forall i\neq I\). In the limit of \(\beta_{\text{eff}} \rightarrow \infty\) the sum can be approximated by a single term, namely \(\sum_{i=1}^N\exp(\beta p_i) \approx \exp(\beta p_I)\). One can then write:
\begin{equation}
    p_i' \approx \frac{\exp(\beta_{\text{eff}} p_i)}{\exp(\beta_{\text{eff}} p_I)} = \exp(\beta_{\text{eff}}(p_i-p_I)) \approx \delta_{iI} \text{ as } \beta_{\text{eff}} \rightarrow \infty\\
\end{equation}
This observation indicates that, at low temperatures, the energy minima of the Modern Hopfield Network are situated close to the stored patterns. The value of $\beta_{\text{c}}$ for which the uniform $\mathbf{p}$-vector becomes unstable in this model can be computed. Due to the high degree of symmetry present in the patterns, arising due to our assumptions, 
N different minima will appear simultaneously when the uniform fixed point loses its stability.

\begin{wrapfigure}{r}{0.45\textwidth}
\vspace{-15pt}
\centering
    \includegraphics[width=0.39\textwidth]{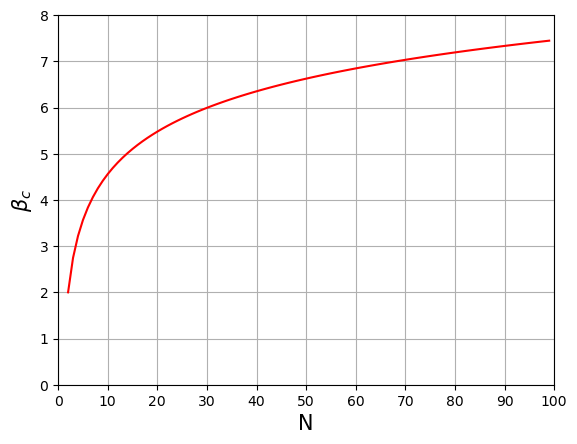}
    \caption{N-dependence of critical inverse temperature $\beta_{\text{c}}$, according to \eqref{eq: PT/ crit equations}.
    }
    \label{fig: PT/N-dependence of beta_c}
\vspace{-55pt}
\end{wrapfigure}
To find at which value of $\beta$ this happens, a cob-web like analysis can be followed \cite{Appendix_StrogatzNonlinDynChaos, RG_book}, such that the following set of equations needs to be solved:
\begin{equation}
\begin{split}
    \mathbf{p^{\prime}} & = \mathbf{p} = \mathbf{p^{\ast}}\\
    \nabla p_1^{\prime} & =  (1,0,...,0)
\end{split}
\end{equation}
These equations can be transformed into
%
\begin{equation}
    \left\{
    \begin{aligned}
    1 & = p_{\text{c}} \left(1 + (N-1)\exp{\frac{1-Np_{\text{c}}}{Np_{\text{c}}(1-p_{\text{c}})}}\right)\\
    \beta_{\text{c}} &= \frac{N-1}{N}\frac{1}{p_{\text{c}}(1-p_{\text{c}})}\end{aligned}\right.
\label{eq: PT/ crit equations}
\end{equation}
with the introduction of the critical values $p_{\text{c}}$, $\beta_{\text{c}}$.\\
Details on how the above equations were derived, are provided in Appendix~\ref{appendix: Critical beta}.
Numerically solving these equations, taking into account that \(p_{\text{c}} \in \left]0, 1\right]\), yields the critical values $p_c$ and $\beta_c$ for any given $N$, as visualized in Fig.~\ref{fig: PT/N-dependence of beta_c}.
\section{Empirical validation}
\begin{figure}[t!]
    \centering
    \begin{subfigure}{0.45\textwidth}
        \centering
        \includegraphics[width=\textwidth]{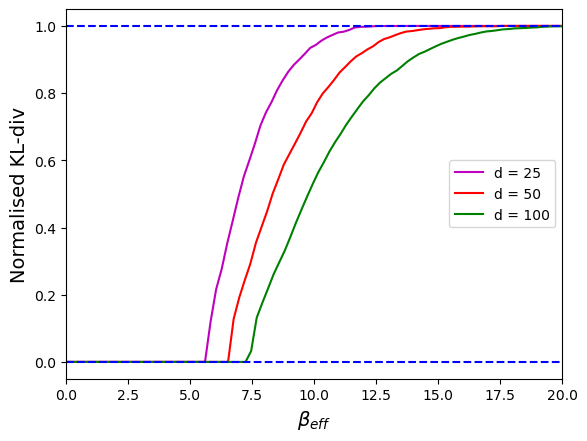}
        \caption{No $\beta$ normalisation}
        \label{fig: EP_MHN/Order param no norm}
    \end{subfigure}
    \hfill
    \begin{subfigure}{0.45\textwidth}
        \centering
        \includegraphics[width=\textwidth]{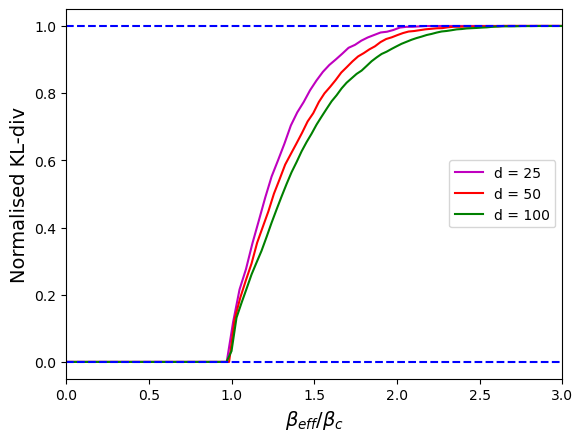}
        \caption{With $\beta$ normalisation}
        \label{fig: EP_MHN/Order param with norm}
    \end{subfigure}
    \caption{$\beta$ dependence of the normalised KL-divergence for different values of $d$. In (b) $\beta$ is normalised with respect to a critical value $\beta_{\text{c}}$ which is obtained analytically by solving \eqref{eq: PT/ crit equations}.}
    \label{fig: EP_MHN/Order param}
\end{figure}

\textbf{Phase transition in MHN with equally spaced and equally normed patterns}\newline
To understand the transition between both limiting cases (uniform vs.~one-hot $\bp$ vector), the Kullback-Leibler divergence ($D_{\text{KL}}$) of $\bp^{\ast}$ w.r.t a uniform distribution is analysed. To make the results independent of the dimension, $D_{\text{KL}}$ is normalised such that the divergence at large $\beta$ is equal to one~\footnote{Mathematically the $D_{\text{KL}}$ is divided by the divergence between a delta distribution and a uniformly distributed vector, i.e. by \(\log(d)\)}.
How $D_{\text{KL}}$ varies as a function of $\beta_{\text{eff}}$ is shown in Fig.~\ref{fig: EP_MHN/Order param}. At the critical point of $\beta_{\text{c}}$ the uniform $\mathbf{p}$-vector becomes unstable. As $\beta$ is further increased, the $\mathbf{p}$-vector becomes increasingly peaked. From these results, it is clear that for an MHN to work as an associative memory model $\beta_{\text{eff}}$ should be chosen larger than $\beta_{\text{c}}$, which is easily achieved by increasing $\beta$.

\textbf{Phase transition in MHN with MNIST patterns} \hfill

\begin{wrapfigure}{r}{0.5\textwidth}
\vspace{-6pt}
\centering
    \includegraphics[width=0.43\textwidth]{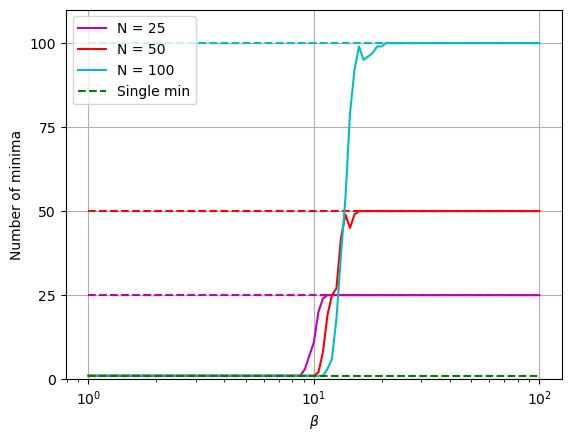}
    \caption{Number of minima in an MHN containing MNIST images as a function of the hyperparameter $\beta$.}
    \label{fig: RealMHN_validation}
\vspace{-10pt}
\end{wrapfigure}
In a final empirical result, we show 
that the presence of a phase transition as described above is not merely a consequence of the strict conditions imposed on the stored patterns \cite{PreciseExponentialStorageCapacity_Lucibello}. To illustrate this, an MHN containing \(N = 25\text{ , } 50\text{ and } 100\) randomly chosen MNIST digits was designed. The number of minima reached starting from noisy versions of the stored patterns was then counted, and 
displayed in Fig.~\ref{fig: RealMHN_validation}. From this figure, it becomes clear that the MHN consisting of freely chosen patterns displays the same kind of phase transition as the simplified network treated in this paper. How the critical parameter $\beta_{\text{c}}$ varies as a function of $N$ also seems to correspond to the observations 
above. Precisely how an effective inverse temperature could be introduced in such a \textit{free} setting is left for future work. 
\section{Conclusion}
An analysis of the Modern Hopfield Network's behavior when tasked with storing equidistant patterns of identical norms provided valuable insights into its core properties.
In particular, we described the presence of a phase transition limiting the regime of high temperatures, at which the energy function displays a single global attractor, to that of low temperatures, at which the energy minima lie infinitesimally close to the stored patterns. 
The critical value at which this transition appeared in the simplified model was analytically obtained. In addition, we discussed the dependence of the dynamics on the \emph{effective inverse temperature},
which also depends on the distribution and norm of the stored patterns. 
As a practical take-away, we have provided insights on how the optimal choice of the hyperparameter $\beta$ for MHNs may vary, depending on the properties of the stored patterns.
\section*{Aknowledgement}
This research was partly funded by the Research Foundation - Flanders (FWO-Vlaanderen) under grant G0C2723N and by the Flemish Government (AI Research Program).
\printbibliography

\newpage

\appendix

\pagenumbering{roman}
\counterwithin{figure}{section}
\renewcommand\thefigure{\thesection\arabic{figure}}

\section{Computing the Critical Inverse Temperature \texorpdfstring{$\beta_{\mathrm{c}}$}{k-e}}
\label{appendix: Critical beta}
\begin{figure}[h!]
    \centering
    \begin{subfigure}{0.48\textwidth}
        \centering
        \includegraphics[width=\textwidth]{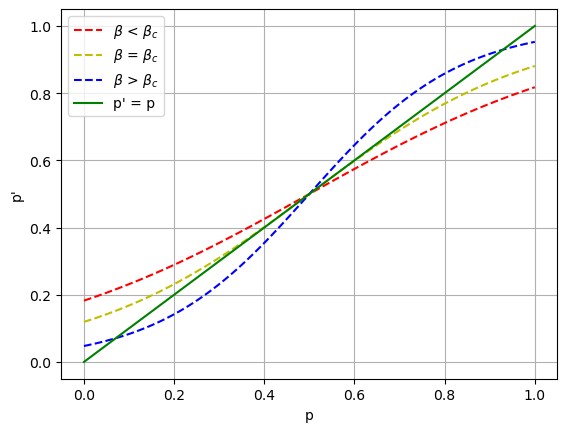}
        \caption{Update rule for cobweb analysis}
        \label{fig: AppendixPT/ 1D update}
    \end{subfigure}
    \hfill
    \begin{subfigure}{0.48\textwidth}
        \centering
        \includegraphics[width=\textwidth]{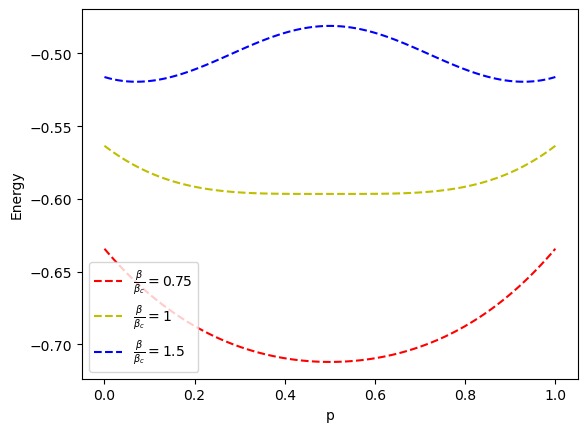}
        \caption{Energy function for different values of $\beta$}
        \label{fig: AppendixPT/ 1D E-func}
    \end{subfigure}
    \caption{In (a), graphical representation of the update rule in one dimension for different values of $\beta$. In (b), the energy function is illustrated for diferent values of $\beta$}
    \label{fig: AppendixPT/ 1D update + CobWeb}
\end{figure}

\begin{figure}
    \centering
    \begin{subfigure}{0.48\textwidth}
        \centering
        \includegraphics[width=\textwidth]{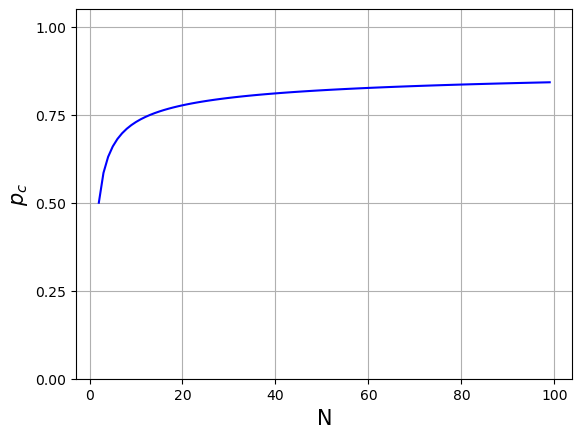}
        \caption{N-dependence of $p_{\text{c}}$}
        \label{fig: AppendixPT/ p_c of N}
    \end{subfigure}
    \hfill
    \begin{subfigure}{0.48\textwidth}
        \centering
        \includegraphics[width=\textwidth]{Associative_Memory_and_Hopfield_Networks_in_2023/Figures/CriticalTemperatureGraph_NoTitle.png}
        \caption{N-dependence of $\beta_{\text{c}}$}
        \label{fig: AppendixPT/ beta_c of N}
    \end{subfigure}
    \caption[N-dependence of critical parameters]{N-dependence of both $p_{\text{c}}$ and $\beta_{\text{c}}$ computed by numerical solving of \eqref{eq: PhaseTransition/ p_c N-dim} and then using \eqref{eq: PhaseTransition/ beta_c N-dim}}
    \label{fig: AppendixPT/ N-dependence of p_c and beta_c}
\end{figure}

In this appendix, the mathematics leading to the expression of the critical parameters are described in more detail. The considered network is again an MHN containing normalised equidistant patterns.\\
To have a better understanding of how the critical values in such a system can be found, it is helpful to first look at the situation with a unique degree of freedom. Due to the normalising condition, this situation is present when \(N=2\), as the system is then entirely characterised by the first component of the $\bm{p}$-vector, namely \(\bm{p} = \left(p,1-p\right)\).\\
In this simple case, the update equation reduces to:\\
\begin{equation}
    p' = f(p) = \frac{\exp(\beta p)}{\exp(\beta p)+\exp(\beta (1-p))} = \frac{\exp(\beta p)}{Z}
\end{equation}
To analyse the fixed point of such a system, it is easiest to do a cobweb analysis \cite{Appendix_StrogatzNonlinDynChaos}. The idea is to simultaneously plot the function \(y = f(p)\) and the first bisector \(y = p\) in the given range of interest, in this case \(p \in \left[0,1\right]\). This is shown for different values of $\beta$ on Fig.~ \ref{fig: AppendixPT/ 1D update}. It should be clear that the intersection of both curves represent the fixed points of the system, described by the equality \(p^{\ast} = f(p^{\ast})\). The stability of a fixed point is then characterised by the slope of the function $f$ at the given fixed point \cite{Appendix_StrogatzNonlinDynChaos}. If the slope is smaller than one, the considered fixed point is stable, while the opposite holds when the slope is larger than one.\\
Two distinct phases of the system are recognisable. If \(\beta<\beta_{\text{c}}\) a unique stable fixed point is found. This fixed point lies precisely in the uniform $\bm{p}$-vector, i.e. in \(p = \frac{1}{2}\). For low values of $\beta$ this uniform fixed point is stable, i.e. \(\frac{\text{d} f}{\text{d}p}\rvert_{p\ast}<1\). At a certain value of $\beta$, which is defined as the critical value $\beta_{\text{c}}$, a pitchfork bifurcation occurs \cite{Appendix_StrogatzNonlinDynChaos}. Two stable fixed points appear, while the uniform fixed point loses its stability. As $\beta$ is increased further, the two stable fixed points go closer and closer to the values of 0 and 1. This highlights that as the temperature decreases, the fixed point of the $\bm{p}$-vector becomes increasingly peaked.\\
This phase transition can also be understood by analysing how the energy function changes for different values of $\beta$. It should be noted that within the made assumptions the energy function introduced by Ramsauer et al. \cite{HopfieldNetworksIsAllYouNeed}, can be rewritten as a function of the $\bm{p}$-vector, i.e. \(E \equiv E(\bm{p},\beta)\). Neglecting the physically irrelevant constants, one finds:\\
\begin{equation}
\begin{split}
E & = -\frac{1}{\beta}\ln\left(\sum_{i=1}^{N} \exp (\beta \bm{x_i^T\xi}) \right) +\frac{1}{2} \bm{\xi^T \xi}\\
& = -\frac{1}{\beta}\ln\left(\sum_{i=1}^{N} \exp (\beta p_i)\prod_{j\neq i}\exp(\beta\cos(\theta) p_j)\right)+\frac{1}{2}\sum_{i=1}^{N}\left(p_i^2+\cos(\theta)\sum_{j \neq i}^{N}p_i p_j\right)\\
\end{split}
\label{eq: AppendixPT/ Energy of p-vec}
\end{equation}\\
\\
This energy function is plotted in 1D for different values of $\beta$ in Fig.~ \ref{fig: AppendixPT/ 1D E-func}. On these plots, the physical cause of the phase transition becomes clear. At high temperatures (\(\beta<\beta_{\text{c}}\)) a single minimum is present residing in the uniform $\bm{p}$-vector. As the temperature is decreased, this minimum becomes flatter and flatter until at some critical value it separates into two minima.\\
The critical value $\beta_{\text{c}}$ can be found by solving the following system:\\
\begin{equation}
\begin{split}
    \left\{
    \begin{aligned}
    p' = f(p) & =  p = p\ast\\
    \frac{\text{d}f}{\text{d}p}\rvert_{p\ast} & = 1
    \end{aligned}\right.
\end{split}
\end{equation}
The first equation represents the fixed point condition. The second equation encompasses the requirement that the uniform fixed point changes stability. This set of equation will not explicitly be solved in this simple illustrative case. Instead, it will first be outlined how this method can be generalised to $N$ dimensions.\\
\\
It should be noted that due to the normalisation condition of the $\bm{p}$-vector, in $N$-dimensions there are $N-1$ degrees of freedom, i.e. \(\bm{p} = (p_1,...,p_{N-1}, 1 - p_1 -... - p_{N-1})\). The update equation for each component is then given by:
\begin{equation}
    p_i' = \frac{\exp(\beta p_i)}{Z}
\end{equation}
Where $Z$ is analogue to the partition function from statistical physics given by \(Z = \sum_{k=1}^N\exp(\beta p_k)\).\\
Notice that, thanks to the symmetry present in the system, it can be understood that when the uniform fixed point loses its stability, $N$ different minima will appear simultaneously, in some sense a pitchfork bifurcation is still present. This originates from the fact that all degrees of freedom can be interchanged freely in the energy function defined by \eqref{eq: AppendixPT/ Energy of p-vec}. This implies that analysing the evolution of one of the degrees of freedom, say $p_1$, should tell us everything there is to know about the system. Imposing the same conditions as previously done in the cob-web analysis gives:
\begin{equation}
\begin{split}
    \left\{
    \begin{aligned}
    \bm{p}' = f(\bm{p}) & =  \bm{p} = \bm{p}\ast\\
    \nabla f(\bm{p})\rvert_{p\ast} & = (1,0,...,0)\\
    \end{aligned}\right.
\end{split}
\label{eq: AppendixPT/ System}
\end{equation}
It should be noted that this simple scheme is not sufficient in systems where such a high degree of symmetry is not present. In practice, one should apply other techniques based on the \textit{Renormalisation Group} \cite{RG_book}. The application of such more intricate methods is however beyond the scope of this article and is left for future work.\\
To solve this system, it proves useful to first consider \(j\neq 1\):\\
\begin{equation}
\begin{split}
    \frac{\partial p_1^{\prime}}{\partial p_j} & = \frac{\partial}{\partial p_j} \frac{\exp\left(\beta_{\text{c}} p_1\right)}{Z}\\
    & = - \frac{\exp\left(\beta_{\text{c}} p_1\right)}{Z^2} \frac{\partial Z}{\partial p_j}\\
    & = - \beta_{\text{c}}\frac{\exp\left(\beta_{\text{c}} p_1\right)}{Z^2}\left[\exp{\beta_{\text{c}} p_j}-\exp{\beta_{\text{c}} p_N}\right]\\
\end{split}
\end{equation}
These simplifications are easily understood when realising that the only two terms in $Z$ depending on $p_j$ are $\exp\left(\beta_{\text{c}} p_j\right)$ and $\exp\left(\beta_{\text{c}} p_N\right)$. Asking that the last line vanishes, as required by \eqref{eq: AppendixPT/ System}, one finds that \(\exp{\beta_{\text{c}} p_j} = \exp{\beta_{\text{c}} p_N}\). Or thus that \(p_j = p_N, \forall j \neq 1\). Hence, as expected, due to the symmetry of the considered model the original $N$ degrees of freedom reduce to a unique degree of freedom which will further simply be denoted as \(p = p_1\). Using the fact that \(\sum_{i=1}^{N}p_i = 1\) it is  then easily found that  \(p_N = \frac{1-p}{N-1}\).\\
The case of \(i = 1\) can then be analysed:
\begin{equation}
\begin{split}
    \frac{\partial p{\prime}}{\partial p} & = \frac{\partial}{\partial p} \frac{\exp\left(\beta_{\text{c}} p\right)}{Z}\\
    & = \frac{\beta_{\text{c}}}{Z^2} \left[ Z \exp\left(\beta_{\text{c}} p\right) - \exp\left(\beta_{\text{c}} p\right) \left(\exp\left(\beta_{\text{c}} p\right) - \exp\left(\beta_{\text{c}} \frac{1-p}{N-1}\right)\right)\right] \\
    & = - \beta_{\text{c}} \left[p-p^2+p\frac{1-p}{N-1}\right] = 1\\
\end{split}
\end{equation}
In the above calculations the fixed point condition was used, i.e. \(p^{\prime} = \frac{\exp\left(\beta p\right)}{Z} = p\).\\
Solving the last line for $\beta_{\text{c}}$:\\
\begin{equation}
    \beta_{c} = \frac{N-1}{N}\frac{1}{p_{c}(1-p_{c})}
\label{eq: PhaseTransition/ beta_c N-dim}
\end{equation}
Finally, by plugging this into the fixed point condition and applying some careful manipulations, one obtains:\\
\begin{equation}
    p_{c} \left(1 + (N-1)\exp{\frac{1-Np_{c}}{Np_{c}(1-p_{c})}}\right) = 1
\label{eq: PhaseTransition/ p_c N-dim}
\end{equation}
This equation is easily solved numerically as it is known that \(p_{c} \in \left]0, 1\right]\), such that the range over which solutions are searched is very limited. Once $p_{c}$ is known, it is sufficient to plug this value back into the equation for $\beta_{c}$ to obtain the critical temperature.\\
Solving this equation repetitively for different values of $N$ the $N$-dependence of $\beta_{\text{c}}$ is obtainable. The $N$-dependence of $p_{\text{c}}$ and $\beta_{\text{c}}$ are shown in Fig.~ \ref{fig: AppendixPT/ N-dependence of p_c and beta_c}. On this graph, a log-like increasing trend for $\beta_{\text{c}}$ can be recognised.
\end{document}